%% file: root.tex
\documentclass[letterpaper, 10 pt, journal, twoside]{ieeetran}
\usepackage{amsmath,amsfonts}
\usepackage{algorithmic}
\usepackage{algorithm}
\usepackage{array}
\usepackage[caption=false,font=normalsize,labelfont=sf,textfont=sf]{subfig}
\usepackage{textcomp}
\usepackage{stfloats}
\usepackage{url}
\usepackage{verbatim}
\usepackage{graphicx}
\usepackage{dsfont}
\usepackage{xcolor}
\usepackage[compress]{cite}
\usepackage{booktabs}
\usepackage{caption}
\usepackage{tabularx}
\usepackage{multirow}
\usepackage{makecell}
\usepackage{float}

\newcommand{\Planner}{\mathcal{P}}
\newcommand{\Step}{f}
\newcommand{\Score}{S}
\newcommand{\Weight}{w}
\newcommand{\Threshold}{\tau}
\newcommand{\Group}{\mathcal{G}}
\newcommand{\Rotation}{R}

\DeclareMathOperator*{\argmax}{argmax}

\newcommand{\new}[1]{\textcolor{black}{#1}}

\begin{document}

\title{Following Is All You Need: Robot Crowd Navigation Using People As Planners}

\author{Yuwen Liao$^{1}$$^{*}$, 
Xinhang Xu$^{1}$$^{*}$, 
Ruofei Bai$^{1}$, 
Yizhuo Yang$^{1}$, \\
Muqing Cao$^{2}$, 
Shenghai Yuan$^{1}$, 
Lihua Xie$^{1}$, \emph{Fellow}, \emph{IEEE}% <-this % stops a space
\thanks{$^{*}$Equal contribution.}%
\thanks{Manuscript received: March 26, 2025; Revised July 17, 2025; Accepted August 3, 2025.}%Use only for final RAL version
\thanks{This paper was recommended for publication by Editor Angelika Peer upon evaluation of the Associate Editor and Reviewers' comments.}%
\thanks{This research is partially supported by the Centre for Advanced Robotics Technology Innovation (CARTIN) funded by the National Research Foundation of Singapore under its Medium-Sized Centre Scheme, and the Agency for Science, Technology and Research (A*STAR) of Singapore under the grant M21K1a0104.}%
\thanks{$^{1}$Authors are with the School of Electrical and Electronic Engineering, 
        Nanyang Technological University, Singapore.
        {\tt\small \{yuwen001, xu0021ng, ruofei001, yizhuo001, shyuan, elhxie\}@ntu.edu.sg}}%
\thanks{$^{2}$Author is with the Robotics Institute, 
        Carnegie Mellon University.
        {\tt\small muqingc@andrew.cmu.edu}}%
\thanks{Digital Object Identifier (DOI): see top of this page.}
}


% Paper headers
\markboth{IEEE Robotics and Automation Letters. Preprint Version. Accepted August, 2025}
{Liao \MakeLowercase{\textit{et al.}}: Following Is All You Need: Robot Crowd Navigation Using People As Planners} 
% Use only for final RAL version


\maketitle

\begin{abstract}
Navigating in crowded environments requires the robot to be equipped with high-level reasoning and planning techniques.
Existing works focus on developing complex and heavyweight planners while ignoring the role of human intelligence.
Since humans are highly capable agents who are also widely available in a crowd navigation setting, we propose an alternative scheme where the robot utilises people as planners to benefit from their effective planning decisions and social behaviours.
% \red{[advantages against existing planner?]}
% \red{[Our method actively and opportunistically identifies suitable human leaders ...]}
Through a set of rule-based evaluations, we identify suitable human leaders who exhibit the potential to guide the robot towards its goal.
Using a simple base planner, the robot follows the selected leader through short-horizon subgoals that are designed to be straightforward to achieve.
We demonstrate through both simulated and real-world experiments that our novel framework generates safe and efficient robot plans compared to existing planners, even without predictive or data-driven modules.
Our method also brings human-like robot behaviours without explicitly defining traffic rules and social norms.
Code will be available at https://github.com/centiLinda/PeopleAsPlanner.git.
\end{abstract}

% Keywords appear just beneath the abstract. Use only for final RAL version. 

\input{1-introduction}

\input{2-related_work}

\input{3-methodology}

\input{4-experiments}

\section{CONCLUSIONS}

This paper presents a novel crowd navigation scheme which utilises people as external intelligent planners.
The more complex planning decisions are handed over to humans so the robot's task is simplified into straightforward subgoal planning.
% only requires a simple base planner to follow the selected leader via straightforward subgoals.
The experiments demonstrate robot behaviours that are safe, efficient, and socially compliant in various crowded environments.
Our proposed framework offers new insights into human-robot relationships in social navigation.
% In a rare scenario when a dense crowd walk against the robot, our method will be reduced to the base planner as no suitable leader can be identified.
% One rare scenario where our method might fall short is when a dense crowd walk against the robot.
% \red{[Not fall short, but reduces to a normal robot navigation method.]}
% It then relies on the base planner's capability to avoid or retreat.
% Future work can be done to infer useful information from other humans who are not leading the robot.
% explore using learning-based methods

\bibliographystyle{IEEEtran}
\bibliography{references}

\end{document}

%% file: 1-introduction.tex
\section{Introduction}

\IEEEPARstart{C}{rowd} navigation is a challenging problem in robotics research as it involves not only static environmental obstacles, but also dynamic agents such as humans. 
% To handle such complex scenarios, modern planners are designed with techniques such as predictive planning, reinforcement learning, and imitation learning \cite{mavrogiannis2023core}.
% Recent research focuses on further improving the performance of these advanced planners through more accurate future estimation \cite{korbmacher2022review} or more diverse and efficient training \cite{sridhar2024nomad, nguyen2023toward}. % zhu2021deep
To handle such complex scenarios, recent research focuses on improving the performance of advanced planners with techniques such as predictive planning and reinforcement learning~\cite{mavrogiannis2023core, korbmacher2022review, nguyen2023toward, guo2025advancements}.%sridhar2024nomad
% All these works aim to develop intelligent and sophisticated robotic systems that can make correct planning decisions, especially in dense crowds and safety-critical scenarios.
These works aim to develop intelligent and sophisticated robotic systems that can make correct planning decisions, especially in dense crowds and safety-critical scenarios.
On the other hand, humans, perhaps the most intelligent agents in the scene, are solving similar navigation problems as the robot.
If we think outside the box, can the robot take advantage of human intelligence to simplify its task?

\begin{figure}
    \centering
    \includegraphics[width=\linewidth]{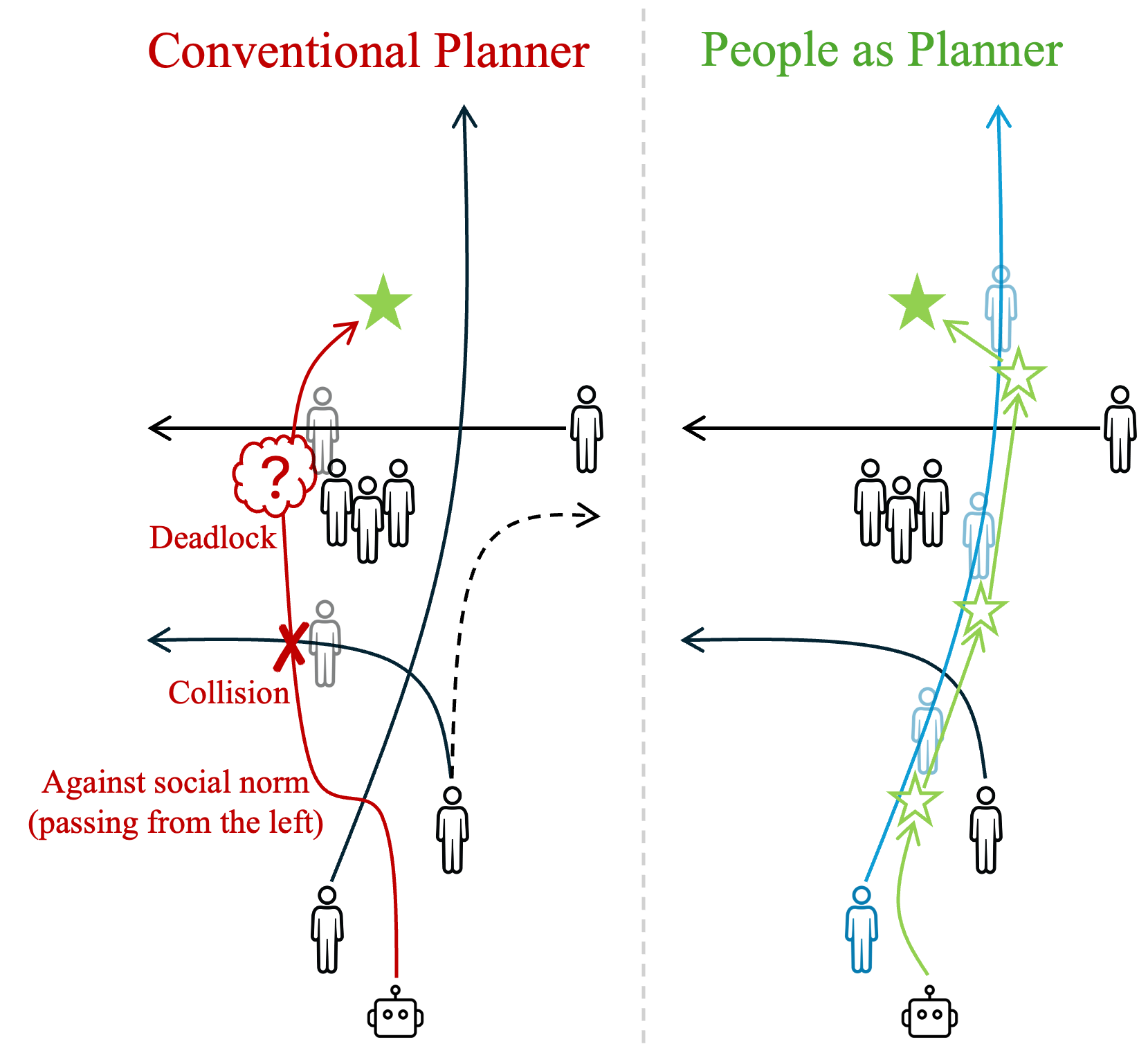}
    % \vspace{-5mm}
    \caption{In a challenging crowd navigation scenario, the robot can easily end up in collisions, deadlocks, and violating implicit social norms. 
    Instead of developing more sophisticated planners, we propose an alternative scheme where the robot follows suitable human leaders to simplify the problem into straightforward subgoal planning.}
    \vspace{-5mm}
    \label{fig:illustration}
\end{figure}

Inspired by how children follow their parents when walking in crowded streets, we find that following trustworthy leaders is an alternative to planning independently, as illustrated in Fig.~\ref{fig:illustration}.
% We can decompose the crowd navigation problem into high-level decisions (e.g. how to cut through a crowd, whether to overtake the person in front) and low-level short-horizon movements.
We can decompose the crowd navigation problem into high-level decisions and low-level short-horizon movements.
The former can be solved using the aforementioned intelligent systems or humans, while the latter only requires simple local planners.
Therefore, we propose to rethink human-robot relationships in crowd navigation by handing over high-level decisions to humans.
% Apart from being dangerous obstacles that the robot needs to avoid, previous works have explored using people as sensors \cite{lewis2009using, afolabi2018people, itkina2022multi, mun2023occlusion} to extend and enrich robot perception.
% In this work, \new{as we assume humans are generally good social navigators, }we use people as intelligent planners that can help the robot solve the more difficult part of the crowd navigation problem.
We use people as intelligent planners that can help the robot solve the more difficult part of the crowd navigation problem.

\begin{figure*}[]
    \centering
    \includegraphics[width=\linewidth]{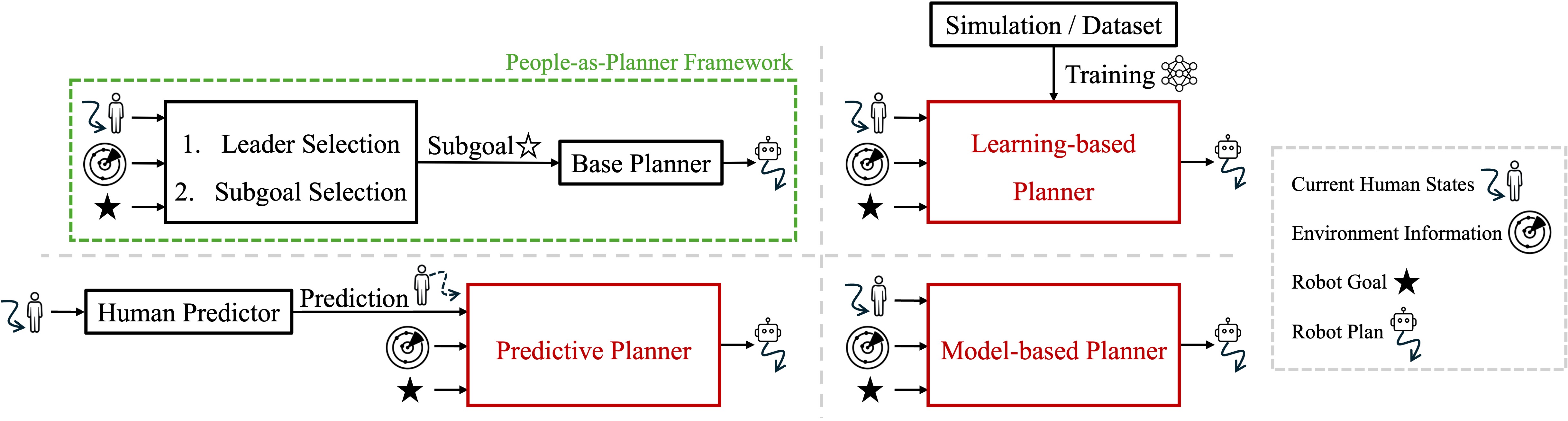}
    % \vspace{-5mm}
    \caption{The proposed People-as-Planner framework, compared with common planning methods in crowd navigation. Utilising humans as high-level planners, we only require the same basic inputs as traditional model-based planners, compared to predictive and learning-based planners which require additional modules. We choose SF as the base planner in experiments.}
    % Traditional model-based planners face difficulties in complex scenarios, while predictive and learning-based planners require additional modules. Our method utilises humans as high-level planners and we choose SF as the base planner in the experiments.}
    \vspace{-3mm}
    \label{fig:framework}
\end{figure*}

Our proposed method consists of two steps, leader selection and subgoal selection.
We start by evaluating nearby humans based on their observed trajectories,
\new{assuming humans are rational and move consistently within a small time window.}
Through a set of rule-based criteria, we aim to identify suitable human leaders who are 1) ahead of and reachable from the robot's current position, 2) heading towards a similar goal at a desired speed.
We then find a subgoal such that it allows the robot to follow the selected leader closely with minimal interruption from other humans.
The subgoals are designed to be easily achievable using a simple base planner and we choose Social Force (SF)~\cite{helbing1995social} in our experiments.
% A subgoal is set near the selected leader and the robot uses a simple Social Force (SF) \cite{helbing1995social} planner to reach the subgoal.
% The subgoal is defined such that it allows the robot to follow the leader closely with minimal interruption from other humans.
The leader selection process is performed continuously so the robot can switch between different humans for maximised efficiency.
% \red{["optimal" is not suitable]}
% The robot will fall back to the base planner if there is no suitable leader to follow.
% This usually happens when the scene is sparse and navigation is straightforward.
We therefore simplify the task from complex crowd navigation to following suitable human leaders using lightweight local planners, which also enables the robot to demonstrate human-like planning behaviours.

Of relevance to our work is the literature on human-following robots. 
% Many of these works study how to maintain a proper distance from humans in different indoor and outdoor scenarios to ensure human comfort \cite{leisiazar2025adapting, lam2010human, morioka2012control}. 
Many of these works study how to maintain a proper distance from humans in different scenarios to ensure human comfort \cite{leisiazar2025adapting, lam2010human}.%morioka2012control 
% Others study how to identify a specific user in crowded environments, or with interrupted observations \cite{antonucci2023humans, algabri2020deep, gupta2016novel}.
Others study how to identify a specific user in complex environments~\cite{antonucci2023humans, algabri2020deep, gupta2016novel}.
In these works, a specific user is predefined for the robot to follow without additional context or objectives.
Our work is fundamentally different as 1) the robot \emph{actively} chooses a suitable leader to follow, 2) by following humans, we effectively address the challenge of crowd navigation.

Our contributions are threefold. 
First, we propose a novel crowd navigation framework that utilises human intelligence to decompose the original problem into straightforward subgoal planning.
Second, we design a set of evaluation processes to find subgoals near suitable human leaders, which can be effectively reached using a simple base planner.
% First, we design an evaluation criteria to select suitable and trustworthy leaders from nearby humans. 
% Second, we propose a leader-following navigation algorithm that decomposes crowd navigation into simpler planning tasks, which can be effectively solved by existing local planners.
Third, we demonstrate through experiments that the proposed People-as-Planner scheme contributes to safe and efficient robot behaviours in crowded and safety-critical scenarios.
% planner & dynamics agnostic plug-and-play module

% assume known topology, a common assumption in recent VLM landmark-based navigation
% not the focus for "Crowd Navigation"

%% file: 2-related_work.tex
\section{Related Work}

\subsection{Social Robot Navigation}
Robot navigation in crowds, or social navigation, has been studied for decades \cite{mavrogiannis2023core, hoy2015algorithms}.
% Traditional model-based methods such as Dynamic Window Approach \cite{fox1997dynamic}, Social Force \cite{helbing1995social}, Optimal Reciprocal Collision Avoidance \cite{van2011reciprocal} are widely adopted in simulations.
Traditional model-based methods~\cite{fox1997dynamic, helbing1995social, van2011reciprocal} have been widely adopted in simulations.
However, they are less suitable for real-world complex scenarios as they are short-sighted and tend to over-simplify human behaviours \cite{trautman2010unfreezing}.
Predictive planning enables the robot to look into the future and avoid potential collisions.
They rely heavily on the performance of prediction modules, as inaccurate predictions may lead to accidents and uncertain predictions result in over-conservative behaviours~\cite{ryu2024integrating, poddar2023crowd}.
End-to-end learning-based methods train robot policies using large-scale demonstrations~\cite{xie2023drl, yang2023st, karnan2022voila}.%chen2019crowd
The performance of the trained networks relies on the accuracy of the simulation platform, or the quality of the collected data~\cite{zhu2021deep}.
% Reinforcement learning develops a value function through trial and error in crowd simulators \cite{xie2023drl, chen2019crowd, liu2021decentralized, yang2023st}.
% To achieve more natural and human-like behaviours, imitation learning uses demonstrations and feedback collected from human users to guide the training process \cite{karnan2022voila, qin2021deep, nguyen2023toward}.
\new{Velocity-field-based methods use crowd motion patterns to inform implicit social norms. 
Their experiment scenarios are usually simple crowd flows in simulations or indoor settings~\cite{dugas2022flowbot, cai2023sampling, zhu2024fast, zhu2023cliff}, 
as complex human movements in the real world are difficult to describe using such homogenised fields.}
% These existing methods, regardless of the complexity, require the robot to plan its path independently. 
Existing methods require the robot to plan its path independently. 
We propose a different approach where the more complex navigation decisions are solved by humans, 
% and the robot is given a simpler task to reach nearby subgoals.
and the robot simply needs to reach nearby subgoals.

\subsection{Human-Following Robot}
Examples of social navigation scenarios include nursing and warehouse assistance, where robots are tasked to follow a target user.
% In applications such as patient monitoring and warehouse assistance, robots are tasked to follow a target user.
Such systems generally consist of modules including human detection, human tracking, and path planning \cite{islam2019person}, which have been extensively studied.
% Recent works focus on more specific tasks in human-following.
% \cite{leisiazar2025adapting} proposes frontal following where the robot needs to predict the human's future trajectories and stay ahead of the human.
% \cite{antonucci2023humans} and \cite{algabri2020deep} use camera-Lidar fusion to track the same person under illumination changes and occlusions.
% \cite{scheidemann2024obstacle} additionally incorporates an RF transponder to identify the target human.
% There are also works that explore new scenarios such as underwater~\cite{islam2018dynamic} and aerial \cite{chen2023quadcopter}.
Recent works focus on more specific tasks such as frontal following~\cite{leisiazar2025adapting}, sensor fusion~\cite{antonucci2023humans, algabri2020deep, scheidemann2024obstacle}, and new application scenarios~\cite{chen2023quadcopter}. %islam2018dynamic
\new{In terms of following strategy, some works replicate human trajectories exactly~\cite{antonucci2023humans, algabri2020deep}, while others define subgoals near the leader~\cite{scheidemann2024obstacle, islam2019person}. We follow the latter as it reduces the perceived intentionality of tracking and results in more natural behaviours.}
We direct the reader to \cite{islam2019person} for a comprehensive review.
Existing human-following methods assign a hypothetical task to the robot to follow a predefined human target.
In our work, the robot's objective is instead to navigate to its destination by actively selecting and following human leaders.

\subsection{Human-Robot Relationships}
% Human-robot interactions involve complex relationships between humans and robots.
% Apart from being dynamic obstacles that the robot needs to avoid, humans often play the role of commanders that perform verbal instructions and teleoperations, or collaborators in tasks such as collaborative assembly and object handover \cite{li2023proactive, selvaggio2021autonomy}.
% For the robot to benefit from human intelligence, \cite{lewis2009using} first propose the concept of using people as sensors.
As the robot navigates in crowds, there exist complex relationships between humans and the robot.
Apart from considering humans as dynamic obstacles, 
% that the robot needs to avoid, 
\cite{lewis2009using} first proposes the concept of using people as sensors to benefit from human intelligence.
% This line of work infers the positions of occluded obstacles from the avoidance behaviours of humans or human drivers \cite{afolabi2018people, itkina2022multi, mun2023occlusion, qiu2024inferring}.
This line of work infers the positions of occluded obstacles from observed human behaviours~\cite{afolabi2018people, itkina2022multi, mun2023occlusion, qiu2024inferring}.%qiu2024inferring
% Inspired by these works, we propose to use people as planners to help the robot make informed navigation decisions.
Inspired by these works, we propose to use people as planners in crowd navigation.
% There are a few works that also consider human-following behaviours in robot planning but with different focuses.
There are a few works that also follow human behaviours in robot planning but with different focuses.
\cite{7759200} and \cite{chen2021unified} consider human-following as one of the policies when generating trajectory samples in a traditional sampling-based planner.
\cite{antonucci2023humans} and \cite{buckeridge2023mapless} utilise human guidance for pathfinding in mapless environments with low human density.
\cite{8793608} imitates pedestrian behaviours on a narrow sidewalk with limited forms of interactions between humans and the robot.
In comparison, our proposed People-as-Planner framework can be applied to various crowded navigation scenarios and is more suitable for real-world robot deployment.

% \cite{antonucci2023humans} describes a similar scenario where the robot follows a human to navigate in an unknown environment.
% However, they simply set one leader at the initialization phase and assume this human can bring the robot to the destination.
% Their research interest lies in human tracking and trajectory recovery, which is fundamentally different from ours.

%% file: 3-methodology.tex
\section{Methodology}

\noindent\textbf{Problem Definition:} 
The state of an agent 
% \red{[Use robot here?]} 
consists of its position and velocity $\boldsymbol{x} = [\boldsymbol{p}^{\top}, \boldsymbol{v}^{\top}]^\top \in \mathbb{R}^{4}$.
We use subscripts $H, R$ to represent human and robot, and superscript $t$ to represent timesteps.
We assign a unique ID $H_{i}$ where $i \in \{1, \dots, n\}$ to $n$ humans that are within the robot's observable range.
The robot's goal position is defined as $\boldsymbol{p}_{g} \in \mathds{R}^{2}$.
% The historical trajectories of all humans can be represented by $\boldsymbol{X}_{H}^{0:t}$.
$\boldsymbol{X}_{H}^{0:t}$ represents the historical trajectories of all humans.
% The robot's goal is defined as $\boldsymbol{g}_{R} = [\boldsymbol{p_{g}}, \boldsymbol{v_{g}}]^\top \in \mathds{R}^{4}$ with both position and velocity.
% The environmental obstacles within the robot's observable range are represented by $\mathcal{O}$.
$\mathcal{O}$ represents the environmental obstacles within the robot's observable range.
In a crowd navigation setting, the robot plan is generated by a planner $\Planner$, denoted as
\begin{equation}
    \boldsymbol{x}_{R}^{t+1} \gets \Planner(\boldsymbol{x}_{R}^{t}, \boldsymbol{x}_{H}^{t}, \mathcal{O}, \boldsymbol{p}_{g}).
\end{equation}
% The planner $\Planner$ replans at every step until the robot reaches its goal.
The planner $\Planner$ replans at every step until goal is reached.

\noindent\textbf{Overview:} 
In our proposed scheme, we first identify a human leader $H_{L}^{t}$ 
% (\textcolor{blue}{MQ: can drop the superscript t})
and subsequently define a subgoal $\boldsymbol{p}_{g}^{t}$:
\begin{subequations}
    
\begin{equation}
    H_{L}^{t} = \Step_{\text{leader}} (\boldsymbol{x}_{R}^{t}, \boldsymbol{X}_{H}^{0:t}, \mathcal{O}, \boldsymbol{p}_{g}), 
\end{equation}
\begin{equation}
    \boldsymbol{p}_{g}^{t} = \Step_{\text{subgoal}} (\boldsymbol{x}_{H_{L}^{t}}^{t}, \boldsymbol{x}_{H}^{t}),
\end{equation}
\end{subequations}
where $\Step_{\text{leader}}(\cdot)$ and $\Step_{\text{subgoal}}(\cdot)$ are the leader and subgoal selection processes that will be detailed in Sec.~\ref{sec:selection} and Sec.~\ref{sec:following}.
Robot plan is then generated by base planner $\Planner$:
\begin{equation}
    \boldsymbol{x}_{R}^{t+1} \gets \Planner(\boldsymbol{x}_{R}^{t}, \boldsymbol{x}_{H}^{t}, \mathcal{O}, \boldsymbol{p}_{g}^{t}),
\end{equation}
% \red{[Eq. 3 duplicate with Eq. 1]}
and the process repeats until goal is reached.
We assume that the robot dynamics is incorporated into the planner and our method is planner-agnostic.
Since each subgoal $\boldsymbol{p}_{g}^{t}$ is much easier to reach than the original goal $\boldsymbol{p}_{g}$, we can use a simple local planner as the base planner.
The overall framework is illustrated in Fig.~\ref{fig:framework}.

We will now introduce the two major components of our proposed method: 1) Leader Selection which involves group identification, reachability check, and individual evaluation, 2) Subgoal Selection which defines a straightforward subgoal near the selected leader and adjusts the following speed.
% We use $\Threshold$ and $\Weight$ to represent thresholds and weights \red{[whose thresholds? You can introduce them later when they are used]}, which are tunable parameters.
Time superscripts are omitted when no confusion is aroused.

\subsection{Leader Selection}
\label{sec:selection}
% \red{[The leader selection involves several steps, including human clustering, reachability checking, and individual evaluation.]}
Before evaluating individual humans, we first identify humans that walk closely together as groups.
Two Human $i$ and Human $j$ are assigned the same group if
\[\Vert \overrightarrow{\boldsymbol{p}_{j} \boldsymbol{p}_{i}} \Vert 
\le \Threshold_{\text{group\_dis}}
, \quad
\Vert \overrightarrow{\boldsymbol{v}_{j} \boldsymbol{v}_{i}} \Vert 
\le \Threshold_{\text{group\_vel}},
\]
% \begin{equation}
% \begin{aligned}
%     \| \overrightarrow{\boldsymbol{p}_{j} \boldsymbol{p}_{i}} \| 
%     &\leq \Threshold_{\text{group\_dis}} \quad \& \\
%     \| \overrightarrow{\boldsymbol{v}_{j} \boldsymbol{v}_{i}} \| 
%     &\leq \Threshold_{\text{group\_vel}},
% \end{aligned}
% \end{equation}
% \begin{equation}
% \begin{aligned}
%     \| \boldsymbol{p}_{i} - \boldsymbol{p}_{j} \| &\leq \Threshold_{\text{group\_dis}} \quad \& \\
%     \| \boldsymbol{v}_{i} - \boldsymbol{v}_{j} \| &\leq \Threshold_{\text{group\_vel}},
% \end{aligned}
% \end{equation}
where $\Threshold_{\text{group\_dis}}$ and $\Threshold_{\text{group\_vel}}$ are thresholds for distance and velocity difference.
We denote the obtained group information as $\Group$ which includes a list of all groups.
Treating groups as a whole can prevent the robot from cutting in between humans that walk together, which respects the social norm.

\begin{figure}[h]
    \centering
    \includegraphics[width=\linewidth]{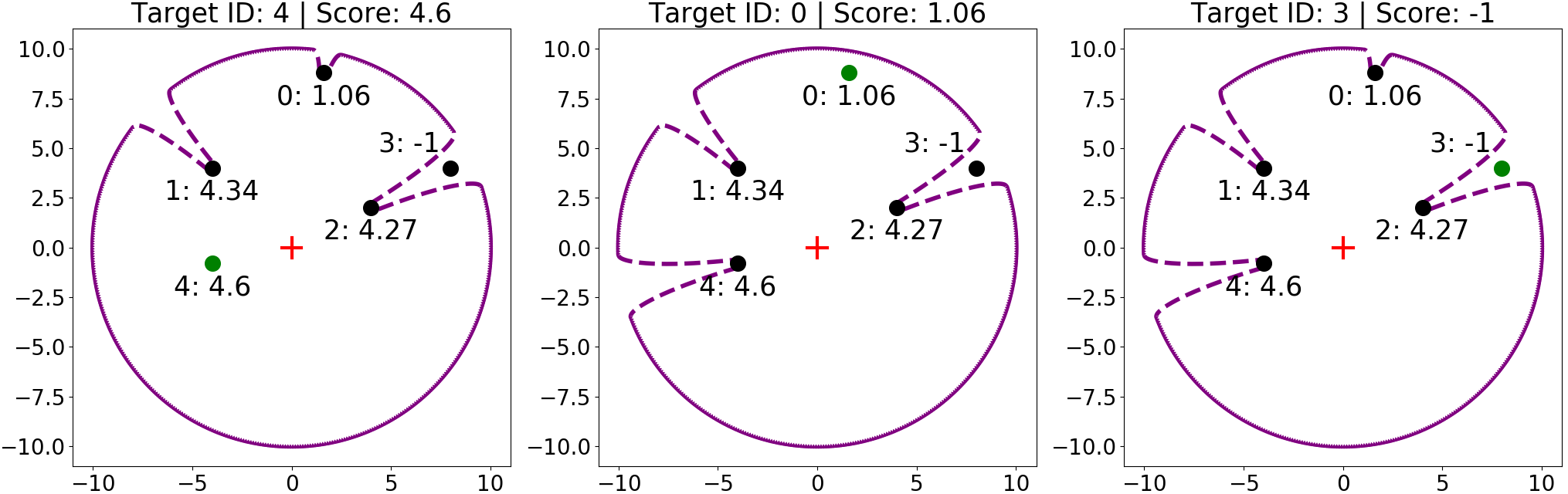}
    % \vspace{-5mm}
    \caption{An example of reachability check from robot (red cross) to three candidate leaders (green dot). The purple curves enclose the visible region. Human 4 has a larger score than Human 0 for being closer to the robot. Human 3 has a negative score as it is blocked by Human 2.}
    % \vspace{-3mm}
    \label{fig:reachability}
\end{figure}

Next, we evaluate the reachability of each human.
As we aim to use simple base planners, the selected leader needs to be directly accessible without the need for complex obstacle avoidance.
Spatio-temporal planning \cite{ding2019safe, zhang2020trajectory} is a feasible method, but at a high computational cost with prediction uncertainties.
Instead, we adopt Line-of-Sight (LoS) distance metric proposed by \cite{bai2025realm} to define the reachability.
As illustrated in Fig.~\ref{fig:reachability}, LoS measures the distance from each human to the boundary of the robot's visible region constructed from LiDAR scans.
% Spatio-temporal planning \cite{ding2019safe, zhang2020trajectory} is also feasible, but at a higher computational cost.
% This reachability can be checked using spatio-temporal planning \cite{ding2019safe, zhang2020trajectory}.
% In actual implementations, we find that it is sufficient to evaluate visibility, as a human that is not visually blocked by any other humans is likely to be reachable.
% \red{[the reason here is not convincing. You can say we use visibility to define which human is reachable, and then mention spatial-termpoal planning as an alternative but is much more time-consuming.]}
% Therefore, we incorporate xx $\Step_{\text{vis}}$ proposed by \red{[Add cite here]} to calculate the visibility of each human, 
We represent the reachability score as $\Step_{\text{reach}}$ and only consider Human $i$ as a potential leader if
\begin{equation}
    \Step_{\text{reach}} (H_{i}, \boldsymbol{x}_{R}^{t}, \boldsymbol{x}_{H}^{t}, \mathcal{O}, \Group) \geq \Threshold_{\text{reach}},
\end{equation}
where $\Threshold_{\text{reach}}$ is the reachability threshold.
% An example of the reachability check is shown in Fig.~\ref{fig:reachability}.
% Details of the algorithm can be found in \cite{bai2025realmrealtimelineofsightmaintenance}.

We then evaluate each qualified human $i$ using 3 criteria: 1) average heading, 2) average speed, 3) relative position between the human, the robot, and the robot's goal.
% \red{[Given equation of $\Score_{i}$ here, and then explain other notations step by step]}
Based on $T$ previous steps, we first evaluate if the human's average heading is towards the robot's goal $\boldsymbol{p}_{g}$:
\begin{equation}
\begin{aligned}
    \Score_{\text{head}} =
    \begin{cases} 
    \frac{\tilde{\boldsymbol{v}}_{i} \cdot 
    \overrightarrow{\boldsymbol{p}_{i} \boldsymbol{p}_{g}}}
    {\| \tilde{\boldsymbol{v}}_{i} \|
    \| \overrightarrow{\boldsymbol{p}_{i} \boldsymbol{p}_{g}} \|}
    &\text{if }
    \frac{\tilde{\boldsymbol{v}}_{i} \cdot 
    \overrightarrow{\boldsymbol{p}_{i} \boldsymbol{p}_{g}}}
    {\| \tilde{\boldsymbol{v}}_{i} \|
    \| \overrightarrow{\boldsymbol{p}_{i} \boldsymbol{p}_{g}} \|}
    \geq \cos \frac{\pi}{4}, \\
    -1, 
    &\text{otherwise}.
    \end{cases}
\end{aligned}
\end{equation}
% where $\tilde{\boldsymbol{v}}_{i} = \frac{\sum_{k=t-T+1}^{t} \boldsymbol{v}_{i}^{k} }{T}$.
where $\tilde{\boldsymbol{v}}_{i} = \frac{1}{T} \sum_{k=t-T+1}^{t} \boldsymbol{v}_{i}^{k}$.
\new{We choose not to predict future headings because existing prediction methods may overlook corner cases~\cite{madjid2025trajectory} and increase computational costs.}

Second, we compare the human's average speed to the robot's ideal speed $v_{\text{pref}}$:
\begin{equation}
\begin{aligned}
    \Score_{\text{vel}} =
    \begin{cases} 
    \frac{\tilde{v}_{i} - v_{\text{pref}}}{v_{\text{pref}}},
    &\text{if }
    \tilde{v}_{i} < v_{\text{pref}}, \\
    \max\left(0, 1 - \frac{\tilde{v}_{i} - v_{\text{pref}}}{v_{\text{pref}}}\right), 
    &\text{otherwise}.
    \end{cases}
\end{aligned}
\end{equation}
% where $\tilde{v}_{i} = \frac{\sum_{k=t-T+1}^{t} \| \boldsymbol{v}_{i}^{k} \|}{T}$.
where $\tilde{v}_{i} = \frac{1}{T} \sum_{k=t-T+1}^{t} \| \boldsymbol{v}_{i}^{k} \|$.
Humans walking at the ideal speed are given the highest score and the score decreases as human speed exceeds the ideal speed.
We penalize slower speeds to encourage active progression.

Third, we compare relative positions to identify humans between robot and its goal, 
% favouring those closer to the robot:
favouring the closer ones:
\begin{equation}
\begin{aligned}
    \Score_{\text{pos}} =
    \begin{cases} 
    \max\left(0, 1 - \frac{\| \overrightarrow{\boldsymbol{p}_{R} \boldsymbol{p}_{i}} \|}{r}\right),
    &\text{if }
    \frac{\overrightarrow{\boldsymbol{p}_{R} \boldsymbol{p}_{i}} \cdot 
    \overrightarrow{\boldsymbol{p}_{R} \boldsymbol{p}_{g}}}
    {\| \overrightarrow{\boldsymbol{p}_{R} \boldsymbol{p}_{i}} \|
    \| \overrightarrow{\boldsymbol{p}_{R} \boldsymbol{p}_{g}} \|}
    > 0, \\
    -1, 
    &\text{otherwise},
    \end{cases}
\end{aligned}
\end{equation}
where $r$ is the robot's observable range. 

We can now calculate a weighted score $\Score_{i}$ by
\begin{equation}
    \Score_{i} = \Weight_{\text{head}} \Score_{\text{head}} + \Weight_{\text{vel}} \Score_{\text{vel}} +\Weight_{\text{pos}} \Score_{\text{pos}},
\end{equation}
where $\Weight_{\text{head}}$, $\Weight_{\text{vel}}$, and $\Weight_{\text{pos}}$ are the weights for each score component.
If $\Score_{i}$ exceeds a threshold $\Threshold_{\text{leader}}$, Human $i$ becomes a leader candidate who can lead the robot towards its goal at a desired speed.
We add an adjustment term to the previous leader $\Score_{H_{L}^{t-1}}$ to avoid fluctuation between candidates with similar scores.
The candidate with the highest score is selected as $H_{L}^{t}$.

\subsection{Subgoal Selection}
\label{sec:following}
We aim to define a subgoal $\boldsymbol{p}_{g}^{t}$ that allows the robot to follow $H_{L}^{t}$ through a straightforward path.
If $H_{L}^{t}$ belongs to any group in the group list $\Group$, the group member closest to the robot is used to replace $H_{L}^{t}$ so that the robot will not attempt to interrupt the group.

We first sample a set of position candidates $\boldsymbol{p}_{m}$ between the robot and $H_{L}^{t}$, defined as
\begin{equation}
        \boldsymbol{p}_{m} = 
    \boldsymbol{p}_{H_{L}^{t}} - 
    \Rotation (\theta_{m}) \cdot 
    \overrightarrow{\boldsymbol{p}_{R} \boldsymbol{p}_{H_{L}^{t}}} \cdot
    \frac{d}
    {\| \overrightarrow{\boldsymbol{p}_{R} \boldsymbol{p}_{H_{L}^{t}}} \|},
\end{equation}
where $\theta_{m}\in \{-\frac{\pi}{4} + m\Delta \theta ~\vert~ 
m = 0, 1, \cdots, \lceil\frac{\pi}{2 \Delta \theta}\rceil
\}$; 
$d$ is the safe distance from humans;
and
\begin{equation}
    \Rotation (\theta_m) = 
    \begin{bmatrix}
    \cos(\theta_m) & -\sin(\theta_m) \\
    \sin(\theta_m) & \cos(\theta_m)
    \end{bmatrix}.
\end{equation}
% \red{[$H_{L}^{t}$ is an index, or the vector of human position?]}
% \begin{subequations}
% \begin{equation}
%     \theta_{m} = m\Delta\theta, \quad m \in \left\{-\frac{45^\circ}{\Delta\theta}, \dots, -1, 0, 1, \dots, \frac{45^\circ}{\Delta\theta}\right\},
% \notag
% \end{equation}
% % rotation matrix
% \begin{equation}
% \begin{aligned}
%     \text{Let } \Rotation (\theta) = 
%     \begin{bmatrix}
%     \cos(\theta) & -\sin(\theta) \\
%     \sin(\theta) & \cos(\theta)
%     \end{bmatrix},
% \notag
% \end{aligned}
% \end{equation}
% \begin{equation}
%     \boldsymbol{p}_{m} = 
%     \boldsymbol{p}_{H_{L}^{t}} - 
%     \Rotation (\theta_{m}) \cdot 
%     \overrightarrow{\boldsymbol{p}_{R} \boldsymbol{p}_{H_{L}^{t}}} \cdot
%     \frac{d}
%     {\| \overrightarrow{\boldsymbol{p}_{R} \boldsymbol{p}_{H_{L}^{t}}} \|},
% \end{equation}
% \end{subequations}
We select the position furthest from neighbouring humans:
\begin{equation}
    \boldsymbol{p}_{g}^{t} = \argmax_{\boldsymbol{p}_{m}}
    \left(
    \min_{H_{i} \neq H_{L}^{t}} \| \overrightarrow{\boldsymbol{p}_{m} \boldsymbol{p}_{H_{i}}} \|
    \right),
\end{equation}
so to minimize the possibility of collision avoidance.

We adjust the robot's speed in the planner $\Planner$ to encourage the robot to follow the leader closely.
In our experiments, we update the speed limit $v_{\text{max}}$ in SF planner directly:
\begin{equation}
\begin{aligned}
    v_{\text{max}} =
    \begin{cases} 
    \| \boldsymbol{v}_{H_{L}^{t}}\|,
    &\text{if }
    \| \overrightarrow{\boldsymbol{p}_{R} \boldsymbol{p}_{H_{L}^{t}}} \|
    \leq \Threshold_{\text{catchup}}, \\
    v_{\text{catchup}}, 
    &\text{otherwise},
    \end{cases}
\end{aligned}
\end{equation}
where $v_{\text{catchup}}$ is a slightly faster speed when the robot needs to catch up with the leader.
For other planners with different settings, this adjustment can be adapted accordingly.

If no $H_{L}$ is identified in the previous leader selection step, we will set $\boldsymbol{p}_{g}^{t} = \boldsymbol{p}_{g}$ where it will plan towards the robot's goal directly.
We will show in Sec.~\ref{sec:qualitative} that this happens when there are only a few surrounding humans and using a simple SF planner is sufficient.

%% file: 4-experiments.tex
\section{Experiments}

\begin{figure}
    \centering
    \includegraphics[width=\linewidth]{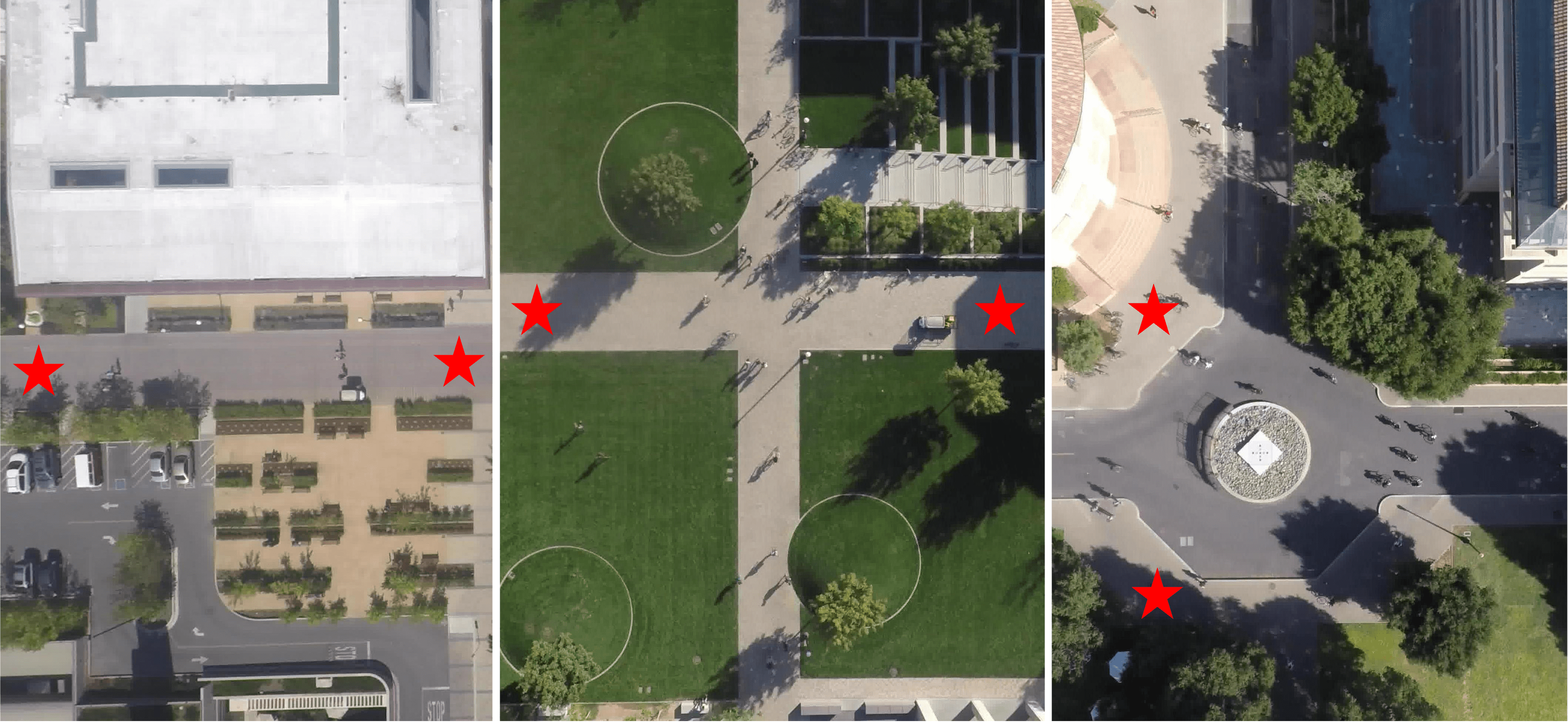}
    % \vspace{-5mm}
    \caption{Three test scenes from the SDD dataset: Promenade, Crossing, and Roundabout. The initial and goal positions are shown in red stars. The direction of travel is interchangeable. }
    % \vspace{-1mm}
    \label{fig:scenes}
\end{figure}

\subsection{Simulation Settings}
\noindent\textbf{Scenarios:}
We first perform simulation experiments with real-world trajectory data from the Stanford Drone Dataset (SDD) \cite{robicquet2016learning}.
SDD was collected mainly from three scenes and we name them Promenade, Crossing, and Roundabout.
As shown in Fig.~\ref{fig:scenes}, the three scenes exhibit increasing levels of difficulty: 1) \textbf{Promenade} has one main road, 2) \textbf{Crossing} has a junction with complex traffic, 3) \textbf{Roundabout} involves challenging street crossing.
% interaction with other road users when crossing the street.
For each scene, we select 10 crowded segments from the longest video.
The scene densities are listed in Table~\ref{tab:quantitative_ours}.
% The segment durations are 45, 45, and 55 seconds respectively for each scene.
% We ensure that the goal is reachable with a slow walking speed of $0.9 \text{m/s}$ (the average walking speed for humans is $1.4 \text{m/s}$ according to \cite{levine1999pace}).
% We allow a longer duration for Roundabout as it involves street crossing which could be time-consuming.
We evaluate our method on all 30 segments and use the densest segment from each scene for comparison study.
As we directly playback the dataset, the humans are non-reactive.
Reactive human behaviours are studied in the real-world experiments in Sec.~\ref{sec:real_world}.

\noindent\textbf{Implementation Details:}
We use a holonomic robot model and the control dynamics is incorporated into the base SF planner.
We set $v_{\text{pref}}$, $r$, $d$ to $1.4 \text{m/s}$~\cite{levine1999pace}, $10 \text{m}$, and $0.8 \text{m}$ respectively.
We set the robot speed limit to $2.0 \text{m/s}$.
The rest of the tunable parameters can be found in our code repository.
We find the set of parameters through manual tuning and we use the same values for all experiments.
% We use the same set of parameters for all experiments.
Since not all baselines distinguish heterogeneous road users, we set the radius of all agents, including robot, humans, bicycles, and cars to 0.5m for fair comparisons.
For collision evaluation, we provide two sets of agent dimensions: 1) all agents have the same radius of 0.5m, 2) bicycles and vehicles are given realistic dimensions of $1.9 \text{m} \times 1 \text{m}$ and $4.5 \text{m} \times 1.9 \text{m}$.
We report results for both settings in Sec.~\ref{sec:quantitative}.
% As SDD dataset consists of heterogeneous road users, we model humans with radius of $0.5m$, bicycles with dimension $1.9 \text{m} \times 1 \text{m}$, and vehicles with dimension $4.5 \text{m} \times 1.9 \text{m}$.
% We model the robot with the same radius as humans and use these dimensions for collision evaluation.
Simulations are conducted at 30Hz on Gazebo platform \cite{koenig2004design} with AMD 5950XT CPU and RTX3090 graphic card.
% Simulations are conducted at 30Hz on Gazebo platform with AMD 5950XT CPU and RTX3090 graphic card.

% Comparisons
\begin{table*}[ht] % Use table* for two-column layout
\centering
\resizebox{\textwidth}{!}{ % Resize to fit text width
\begin{tabular}{|l|ccc|ccc|ccc|}
\hline
\multirow{2}{*}{Method} & \multicolumn{3}{c|}{Promenade} & \multicolumn{3}{c|}{Crossing} & \multicolumn{3}{c|}{Roundabout} \\ 
\cline{2-10}
 & TCC$\downarrow$ & T\textsubscript{avg}(s)$\downarrow$ & D\textsubscript{avg}(m)$\downarrow$ & TCC$\downarrow$ & T\textsubscript{avg}(s)$\downarrow$ & D\textsubscript{avg}(m)$\downarrow$ & TCC$\downarrow$ & T\textsubscript{avg}(s)$\downarrow$ & D\textsubscript{avg}(m)$\downarrow$ \\ 
\hline
SF \cite{helbing1995social} & \underline{2.66} / 50.13 & 50.11 & 62.99 & 125.46 / 234.13 & 39.19 & 54.57 & 13.42 / 77.21 & 22.53 & 31.37 \\ 
DWA \cite{fox1997dynamic} & 44.14 / 114.79 & 48.56 & 67.03 & \underline{8.00} / 182.59 & 29.98 & \underline{42.43} & 29.85 / 141.55 & 27.69 & 39.16 \\ 
ORCA \cite{van2011reciprocal} & 8.00 / \underline{43.11} & 56.62 & 74.81 & \textbf{0.00} / 80.07 & \underline{29.70} & \textbf{41.87} & \underline{2.94} / \underline{53.07} & 28.41 & 35.03 \\ 
Pred2Nav+CV \cite{poddar2023crowd} & 122.71 / 125.68 & 48.45 & 57.34 & 18.90 / \underline{40.68} & 32.64 & 43.50 & 16.27 / 58.12 & 19.84 & 29.37 \\ 
Pred2Nav+SGAN \cite{poddar2023crowd} & 138.36 / 139.65 & \underline{43.92} & \underline{56.06} & 71.47 / 117.06 & 31.98 & 44.22 & 23.87 / 83.66 & \textbf{19.40} & 30.05 \\ 
HEIGHT \cite{liu2024height} & 605.68 / 605.68 & 56.08 & \textbf{43.35} & 171.45 / 270.46 & 29.91 & 44.45 & 12.17 / 71.76 & \underline{19.41} & \textbf{26.17} \\ 
\hline
Ours & \textbf{0.25} / \textbf{1.94} & \textbf{40.39} & 60.48 & \textbf{0.00} / \textbf{0.00} & \textbf{29.40} & 45.10 & \textbf{0.54} / \textbf{5.44} & 19.87 & \underline{29.17} \\ 
\hline
\end{tabular}
}
\caption{Quantitative comparisons on the densest segment from each test scene. 100 repeated experiments are performed for each segment. For TCC, the first and second values (e.g., 0.25 and 1.94 of our method in the Promenade scene) are calculated by setting all agents with the same radius of 0.5m, and setting bicycles and vehicles with realistic dimensions, respectively. TCC is the frames with collisions under 30Hz simulation (e.g., our method leads to collisions in 0.25 frames in the Promenade scene, averaged over 100 runs). The best performance is in \textbf{bold} and the second-best is \underline{underlined}.}
\label{tab:quantitative_compare}
\end{table*}

% 30 scenes
\begin{table*}[t] % Use table* for two-column layout
\centering
\resizebox{\textwidth}{!}{ % Resize to fit text width
\large
\begin{tabular}{|l|c@{\hskip 4pt}c@{\hskip 4pt}c@{\hskip 4pt}c@{\hskip 4pt}c@{\hskip 4pt}c@{\hskip 4pt}c@{\hskip 4pt}|c@{\hskip 4pt}c@{\hskip 4pt}c@{\hskip 4pt}c@{\hskip 4pt}c@{\hskip 4pt}c@{\hskip 4pt}c@{\hskip 4pt}|c@{\hskip 4pt}c@{\hskip 4pt}c@{\hskip 4pt}c@{\hskip 4pt}c@{\hskip 4pt}c@{\hskip 4pt}c@{\hskip 4pt}|}
\hline
\multirow{2}{*}{Seg.} & \multicolumn{7}{c|}{Promenade} & \multicolumn{7}{c|}{Crossing} & \multicolumn{7}{c|}{Roundabout} \\ 
\cline{2-22}
 & TCC$\downarrow$ & T\textsubscript{avg}(s)$\downarrow$ & D\textsubscript{avg}(m)$\downarrow$ & Ped & Others & Total & Density
 & TCC$\downarrow$ & T\textsubscript{avg}(s)$\downarrow$ & D\textsubscript{avg}(m)$\downarrow$ & Ped & Others & Total & Density
 & TCC$\downarrow$ & T\textsubscript{avg}(s)$\downarrow$ & D\textsubscript{avg}(m)$\downarrow$ & Ped & Others & Total & Density\\ 
\hline
1 & 1.94 & 40.39 & 60.48 & 21 & 3 & 24 & high & 0.00 & 29.40 & 45.10 & 21 & 23 & 44 & high & 5.44 & 19.87 & 29.17 & 38 & 51 & 89 & high \\
2 & 0.00 & 40.66 & 60.27 & 24 & 4 & 28 & medium & 0.00 & 28.43 & 44.73 & 18 & 17 & 35 & high & 5.33 & 18.29 & 29.97 & 34 & 49 & 83 & medium \\
3 & 0.00 & 44.40 & 59.25 & 13 & 2 & 15 & low & 0.00 & 28.99 & 45.48 & 33 & 27 & 60 & high & 3.62 & 20.87 & 28.66 & 26 & 43 & 69 & high \\
4 & 0.00 & 37.68 & 57.16 & 23 & 1 & 24 & medium & 0.56 & 30.27 & 50.50 & 27 & 25 & 52 & medium & 0.00 & 16.27 & 28.45 & 25 & 52 & 77 & medium \\
5 & 0.00 & 44.99 & 60.33 & 24 & 2 & 26 & high & 12.41 & 34.04 & 48.09 & 27 & 24 & 51 & medium & 0.00 & 17.70 & 26.60 & 15 & 48 & 63 & low \\
6 & 1.73 & 44.33 & 61.60 & 24 & 3 & 27 & high & 0.00 & 27.26 & 44.17 & 21 & 17 & 38 & medium & 0.00 & 16.70 & 25.99 & 6 & 29 & 35 & low \\
7 & 3.10 & 42.72 & 59.66 & 23 & 4 & 27 & high & 6.70 & 32.45 & 46.93 & 12 & 17 & 29 & medium & 8.38 & 25.08 & 34.54 & 33 & 46 & 79 & high \\
8 & 0.00 & 35.95 & 58.66 & 16 & 3 & 19 & low & 5.00 & 27.82 & 46.57 & 19 & 18 & 37 & medium & 1.75 & 18.53 & 26.81 & 15 & 47 & 62 & low \\
9 & 14.62 & 42.87 & 56.94 & 18 & 4 & 22 & medium & 12.69 & 31.13 & 47.86 & 26 & 22 & 48 & medium & 7.64 & 20.11 & 30.13 & 24 & 51 & 75 & medium \\
10 & 0.00 & 36.88 & 56.09 & 21 & 2 & 23 & medium & 0.00 & 35.50 & 46.92 & 27 & 19 & 46 & low & 7.56 & 17.45 & 27.77 & 28 & 51 & 79 & medium \\
\hline 
Avg & 2.14 & 41.09  & 59.04  & \textbackslash & \textbackslash & \textbackslash & \textbackslash & 3.74  & 30.53  & 46.64  & \textbackslash & \textbackslash & \textbackslash & \textbackslash & 3.97  & 19.09  & 28.80  & \textbackslash & \textbackslash & \textbackslash & \textbackslash \\  
STD &  4.53 &  3.32 &  1.80 & \textbackslash & \textbackslash & \textbackslash & \textbackslash & 5.22 &  2.73 &  1.89 & \textbackslash & \textbackslash & \textbackslash & \textbackslash &  3.37 &  2.58 &  2.46 & \textbackslash & \textbackslash & \textbackslash & \textbackslash \\ 
\hline

\end{tabular}
}
\caption{Quantitative results of our method on the three test scenes with 10 segments (Seg.) each. 100 repeated experiments are performed for each segment. For TCC, we use realistic dimensions for bicycles and vehicles, which corresponds to the second value of TCC in Table~\ref{tab:quantitative_compare}. We count the total number of road users in each segment, including pedestrians, bicycles, and vehicles. We record the crowd density along the robot's path, determined through visual inspection.}
% \vspace{-3mm}
\label{tab:quantitative_ours}
\end{table*}
% Summary
\begin{table}
\centering
% \begin{tabular}{lccccc}
\begin{tabular}{l@{\hskip -6pt}c@{\hskip 4pt}c@{\hskip 4pt}c@{\hskip 4pt}c@{\hskip 4pt}c}
\toprule
Method & \makecell{Current \\ human states} & \makecell{Future \\ human states} & \makecell{Environment \\ information} & Training \\
\midrule
SF \cite{helbing1995social} & \checkmark & \texttimes & \checkmark & \texttimes \\
DWA \cite{fox1997dynamic} & \texttimes & \texttimes & \checkmark & \texttimes \\
ORCA \cite{van2011reciprocal} & \checkmark & \texttimes & \checkmark & \texttimes \\
Pred2Nav+CV \cite{poddar2023crowd} & \checkmark & \checkmark & \texttimes & \texttimes \\
Pred2Nav+SGAN \cite{poddar2023crowd} & \checkmark & \checkmark & \texttimes & \checkmark \\
HEIGHT \cite{liu2024height} & \checkmark & \texttimes & \checkmark & \checkmark \\
\midrule
Ours & \checkmark & \texttimes & \checkmark & \texttimes \\
\bottomrule
\end{tabular}
\caption{Comparison of deployment requirements between baselines and our method.}
% \vspace{-3mm}
\label{tab:compare}
\end{table}

\noindent\textbf{Baselines:}
We compare our method with five crowd navigation methods under different frameworks:
\begin{enumerate}
    \item Social Force (\textbf{SF}) \cite{helbing1995social} calculates attraction from goal and repulsion from obstacles and neighbours.
    \item Dynamic Window Approach (\textbf{DWA}) \cite{fox1997dynamic} searches for the optimal trajectory within a dynamic window that moves towards the goal while avoiding obstacles.
    \item Optimal Reciprocal Collision Avoidance (\textbf{ORCA}) \cite{van2011reciprocal} solves for the optimal action for each agent assuming equal responsibility for avoidance.
    % \item \textbf{SICNav} \cite{samavi2024sicnav} uses bi-level Model Predictive Control (MPC) by predicting humans using ORCA model.
    \item \textbf{Pred2Nav} \cite{poddar2023crowd} incorporates human trajectory predictors into a sampling-based MPC. We compare two predictors implemented by the author, constant velocity (CV) and SGAN \cite{gupta2018social} which is a neural network.
    \item \textbf{HEIGHT} \cite{liu2024height} is an end-to-end reinforcement learning method that considers crowd interactions. We follow the training procedures from the provided repository to train the model for holonomic robots.
\end{enumerate}

\noindent\textbf{Evaluation Metrics:}
We perform 100 repeated runs for each segment and evaluate the average performance to count for the uncertainty of physics engine approximation in simulation.
We mark it as a collision if the robot enters within the agent dimensions.
The following three metrics are used:
\begin{enumerate}
    % \item Success Rate (\textbf{SR}) is the percentage of runs where the robot reaches its goal within the segment duration.
    \item Total Collision Count (\textbf{TCC}) is the number of frames where collision occurs. Experiment continues after a collision so there could be multiple counts in one run.
    \item Average Time (\textbf{T}\textsubscript{\textbf{avg}}) taken to reach the goal.
    \item Average Distance (\textbf{D}\textsubscript{\textbf{avg}}) taken to reach the goal.
    % \item Time-per-step (\textbf{t}\textsubscript{\textbf{step}}) is the average inference time for each step.
\end{enumerate}

\begin{figure*}[ht]
    \centering
    \includegraphics[width=\linewidth]{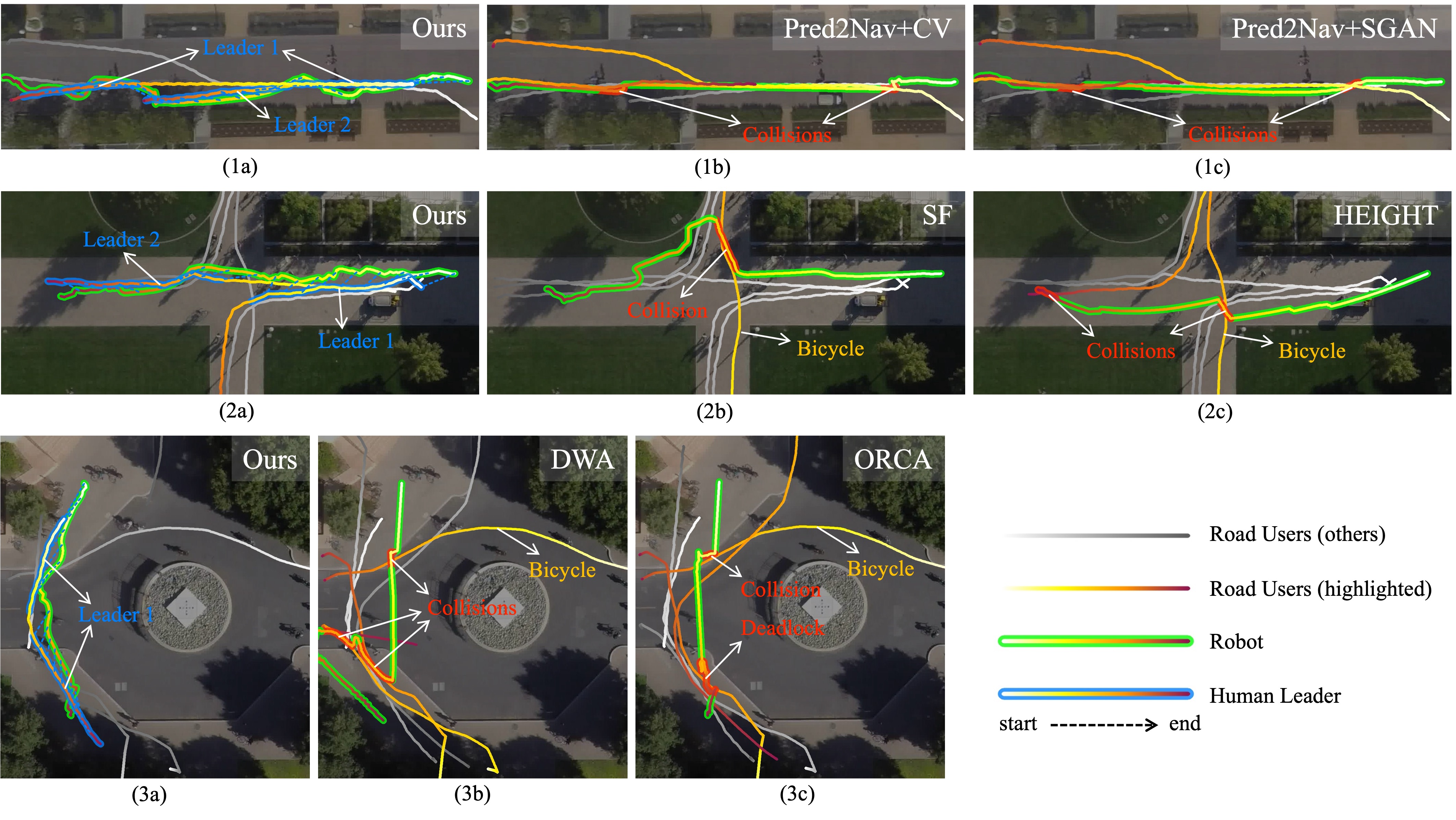}
    % \vspace{-5mm}
    \caption{Qualitative comparisons in simulation experiments. The three rows correspond to the Promenade, Crossing, and Roundabout scenes respectively. The temporal progression is represented by the gradient colour (from light to dark) of each trajectory. The robot and the selected leaders are highlighted by different glow colours. For simplicity, we select a few agents to draw for each scene to highlight the interactions. The full experiment recordings can be found in the supplementary video.}
    \vspace{-3mm}
    \label{fig:qualitative}
\end{figure*}

\subsection{Quantitative Evaluation}
\label{sec:quantitative}
As shown in Table.~\ref{tab:quantitative_compare}, our method shows the best safety awareness with high efficiency across all three scenes.
Our method allows the robot to adjust its speed according to the leader, so it can reach the goal fast even when additional manoeuvres are required for effective collision avoidance.
Notice that although we set the radius for all agents to 0.5m during the experiments, when we calculate collisions using real dimensions for other road users (namely bicycles and vehicles), our method maintains a low TCC, while the values increase dramatically for all other baselines.
This implies that our method can keep a larger distance from dangerous road users without explicitly considering agent types in the inputs, which benefits from the human leader's planning decisions.

In comparison, ORCA tends to get too close to humans since it assumes reciprocal avoidance behaviours, which lead to more collisions than ours.
SF and DWA are less robust in highly dynamic environments, as shown by the inconsistent performance across the three scenes.
For Pred2Nav, although SGAN gives more accurate human predictions on large-scale datasets~\cite{gupta2018social}, it doesn't consistently improve the planning performance.
HEIGHT is less sensitive to collisions because it is not explicitly enforced like the other methods.
It demonstrates biased behaviours such as moving in a curved path, which results in heavy collisions with the wall in the Promenade scene.
We also show that by simplifying complex crowd navigation problems using people as planners, we can achieve the best performance using a simple base planner SF, which is less capable when planning independently.

We summarise the types of information and procedures required to deploy each method in Table.~\ref{tab:compare}.
All methods except DWA require current states of humans to consider human avoidance and interaction explicitly.
All baselines except Pred2Nav require environmental information such as obstacle maps and point clouds for obstacle avoidance.
Although Pred2Nav does not consider static obstacles, none of its TCC reported in Table~\ref{tab:quantitative_compare} comes from collision with walls.
Our method uses point clouds for human reachability evaluations.
In addition, Pred2Nav requires a trajectory prediction module to obtain future human states.
HEIGHT, as well as SGAN used by Pred2Nav, require training on either simulations or large-scale datasets.
We can see that our method only needs the basic inputs to achieve the best overall performance.
Our method is further evaluated over all 30 segments from the three scenes, as shown in Table.~\ref{tab:quantitative_ours}.
We demonstrate strong robustness as the performance is consistent under different traffic conditions and crowd densities.

\subsection{Qualitative Evaluation}
\label{sec:qualitative}
Using people as planners, our method demonstrates efficient and socially-aware behaviours that contribute to the outstanding results presented above.
For the Promenade scene in Fig.~\ref{fig:qualitative}-1a, the robot switched to the second leader when the first leader was temporarily occluded, and switched back when the second leader stopped moving.
In Fig.~\ref{fig:qualitative}-1b, Pred2Nav resulted in collisions twice.
From the sharp turns before the collisions, we can infer that the robot tried to avoid the humans, but it acted too late.
Compared to Fig.~\ref{fig:qualitative}-1c, predicting with SGAN did not bring much benefit.
For the Crossing scene, both SF and HEIGHT resulted in collisions with a fast-moving bicycle, as shown in Fig.~\ref{fig:qualitative}-2b and Fig.~\ref{fig:qualitative}-2c.
Our method, as shown in Fig.~\ref{fig:qualitative}-2a, avoided this encounter by switching between the two leaders.

For the Roundabout scene, DWA first collided with a bicycle, then with two humans walking face-to-face, as shown in Fig.~\ref{fig:qualitative}-3b.
In Fig.~\ref{fig:qualitative}-3c, ORCA only collided with the bicycle, but was forced to stop by the two humans and only continued moving after the humans walked by.
In comparison, as shown in Fig.~\ref{fig:qualitative}-3a, our method followed the leader to cross the road safely, switched to default planner in the middle when the leader was temporarily occluded, and then continued to follow towards the goal.
An interesting observation in this scene is that, although traffic rules are not explicitly defined, our method follows the human to cross the road from the appropriate point on the sidewalk, which avoids a dangerous encounter with the bicycle.
Other baselines, however, move towards the goal directly, which inevitably increases the difficulty of avoiding fast-moving road users on the main road.
This makes our method especially useful in real-world deployments, as it is difficult to define every rule in every scenario, our method can simply follow the surrounding agents to respect the ``unknown rules".

\subsection{Real-World Experiment}
\label{sec:real_world}
% Although SDD dataset provides real-world human data, it cannot reproduce human reactive behaviours towards the robot.
% Therefore, we further demonstrate our method in a 150m crowded campus corridor with over 40 real-world human encounters.
In addition to the performance comparison against baselines using recorded human data, we further demonstrate the effectiveness of our method when interacting with real-world reactive humans.
We conduct the experiment in a 150m crowded campus corridor with over 40 human encounters, which is similar to the previous SDD scenes in terms of environment and human density.
We use a 2D LiDAR to obtain point clouds for reachability check, and use a depth camera for human tracking. 
\new{To ensure human safety and privacy, we reduce the robot speed limit to $1.0 \text{m/s}$, stop the robot when getting too close to pedestrians, and do not collect their identifiable information during experiments.}
% We reduce the robot speed limit to $1.0 \text{m/s}$ to ensure safety, 
% \new{and implement safety measures to stop the robot when getting too close to the pedestrians}.
\begin{figure}[h]
    \centering
    \includegraphics[width=\linewidth]{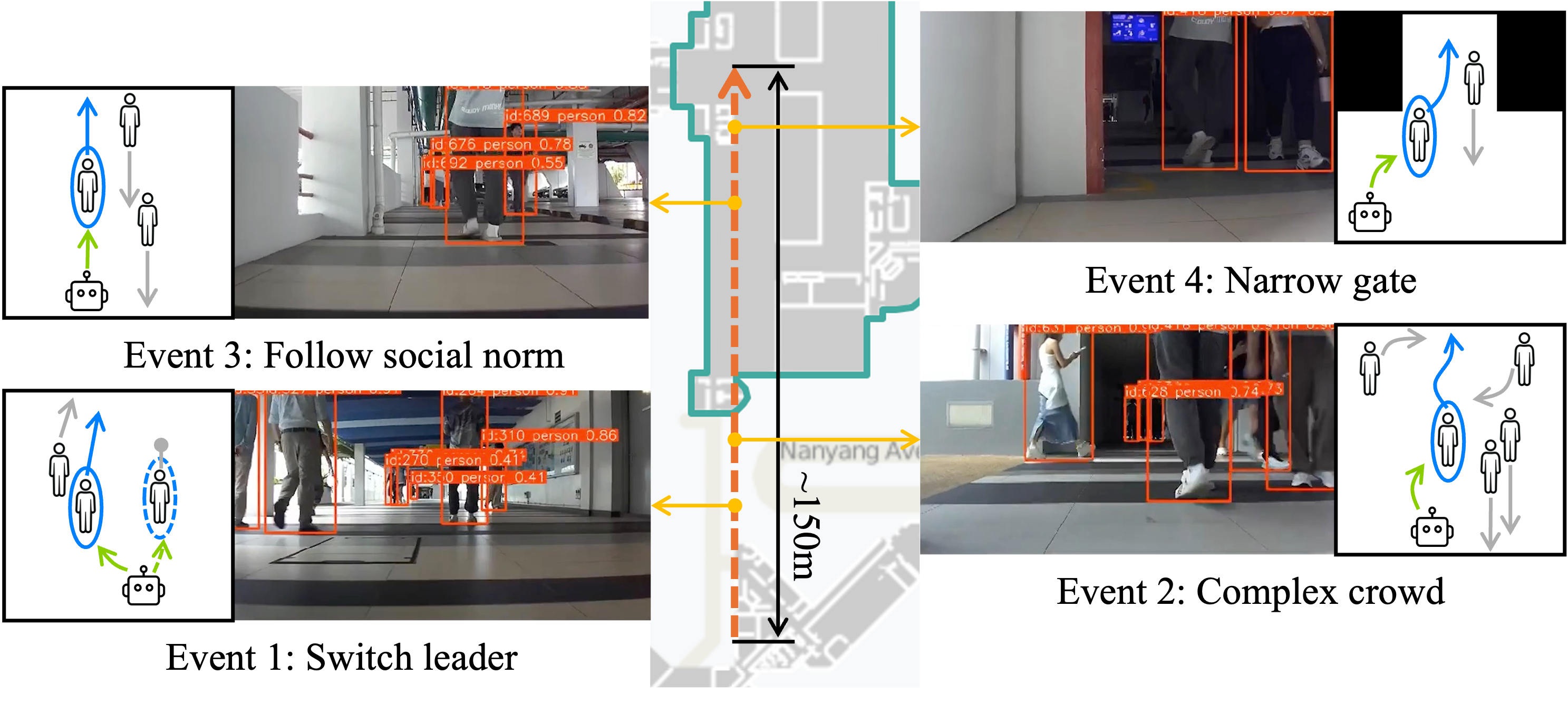}
    % \vspace{-5mm}
    \caption{Events observed in the real-world experiment. Green arrows represent robot subgoal planning. Blue colours represent selected leaders and their movements. Grey arrows represent movements of other humans. The dashed circle in Event 1 represents the previous leader.}
    % \vspace{-3mm}
    \label{fig:realworld}
\end{figure}

In Fig.~\ref{fig:realworld}, we illustrate four events, most of which have been observed in the simulation experiments.
In the first event, the robot switched to a new leader when the previous leader slowed down.
This ensures the robot continues to progress efficiently without being stuck behind static humans.
In the second and fourth events, the robot moved through a complex crowd and a narrow gate without deadlocks.
In the third event, the robot followed the social norm to keep to the left, which avoided encounters with humans walking from the opposite direction.
The robot benefits from the proposed people-as-planner scheme and makes intelligent decisions in real-world crowded environments.
Our method is robust to disturbances such as inaccurate human tracking, hardware delays, etc., and successfully completes the run without collisions.
The complete experiment can be found in the supplementary video.

% % Comparison
% \begin{table*}[t]
% \centering
% \begin{tabular}{lccccc}
% \toprule
% Method & SR(\%)$\uparrow$ & TCC$\downarrow$ & T\textsubscript{avg}(s)$\downarrow$ & D\textsubscript{avg}(m)$\downarrow$ & t\textsubscript{step}(ms)$\downarrow$ \\
% \midrule
% A & 0.46/0.46/0.46 & 0.02/0.02/0.02 & 0.42/0.42/0.42 & 0.00/0.00/0.00 & 0.09/0.09/0.09 \\
% B & 0.46/0.46/0.46 & 0.08/0.08/0.08 & 0.19/0.19/0.19 & 0.06/0.06/0.06 & 0.21/0.21/0.21 \\
% C & 0.56/0.56/0.56 & 0.07/0.07/0.07 & 0.37/0.37/0.37 & 0.01/0.01/0.01 & 0.00/0.00/0.00 \\
% D & 0.56/0.56/0.56 & 0.06/0.06/0.06 & 0.27/0.27/0.27 & 0.05/0.05/0.05 & 0.05/0.05/0.05 \\
% E & 0.56/0.56/0.56 & 0.09/0.09/0.09 & 0.24/0.24/0.24 & 0.04/0.04/0.04 & 0.09/0.09/0.09 \\
% F & 0.24/0.24/0.24 & 0.00/0.00/0.00 & 0.76/0.76/0.76 & 0.00/0.00/0.00 & 0.01/0.01/0.01 \\
% \midrule
% Ours & 0.59/0.59/0.59 & 0.13/0.13/0.13 & 0.21/0.21/0.21 & 0.00/0.00/0.00 & 0.07/0.07/0.07 \\
% \bottomrule
% \end{tabular}
% \caption{Qualitative comparisons from simulation using 40 scenarios selected from the SDD dataset. We report the results for Crossing / Roundabout / Promenade separately. Each scenario is repeated for 100 runs.}
% \label{tab:quantitative_table}
% \end{table*}

\begin{figure}[h]
    \centering
    \includegraphics[width=0.9\linewidth]{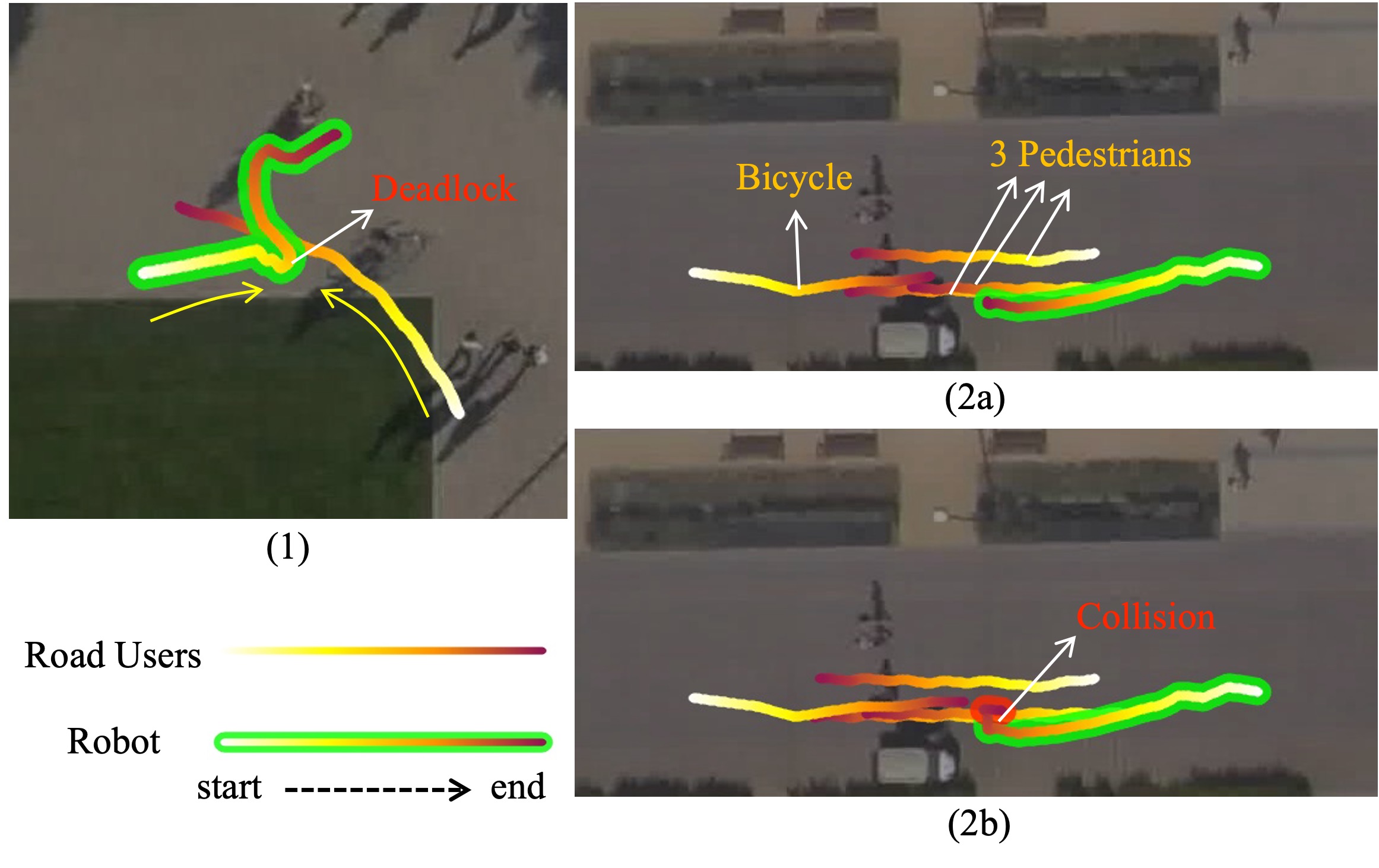}
    % \vspace{-5mm}
    \caption{\new{Failure cases in simulation experiments. In (1), the robot was confused by a human blocking the way and forced into a short deadlock. In (2a), the robot was following a group and did not detect an oncoming bicycle due to occlusion. A dataset labelling error caused the bicycle to unrealistically run over the pedestrians, leading to a sudden collision with the robot in (2b).}}
    % \vspace{-3mm}
    \label{fig:failure_cases}
\end{figure}

\subsection{Limitations}
\label{sec:limitations}
\new{During simulation experiments, we found that our method may fail when interacting with aggressive road users, as shown in Fig.~\ref{fig:failure_cases}.
% Future work can be done to enable faster robot reactions (e.g., temporary high speed to retreat).
Future work could incorporate data-driven policies trained on real-world corner cases.
Some existing works allow the robot to actively influence human behaviours~\cite{dugas2020ian, che2020efficient, angelopoulos2022you}, which may also be useful, especially in difficult scenarios such as moving against a dense crowd.}

\new{The dynamic subgoal setting in the proposed framework may potentially affect trajectory smoothness.
% Nevertheless, as shown in Table.~\ref{tab:acc_jerk}, our method does not bring significant fluctuations~\cite{francis2025principles} compared to the baselines.
% Nevertheless, as shown in Fig.~\ref{fig:acc_jerk}, our method does not bring significant fluctuations~\cite{francis2025principles} compared to the baselines.
In the current experiments, our method does not show significant fluctuations compared to the baselines (comparison results on trajectory smoothness can be found in the supplementary video).
Nevertheless, to improve the smoothness, future work may consider other types of robot dynamics apart from the holonomic model, or incorporate various trajectory smoothing techniques~\cite{ravankar2018path}.}

%% file: root.bbl
\begin{thebibliography}{10}
\providecommand{\url}[1]{#1}
\csname url@samestyle\endcsname
\providecommand{\newblock}{\relax}
\providecommand{\bibinfo}[2]{#2}
\providecommand{\BIBentrySTDinterwordspacing}{\spaceskip=0pt\relax}
\providecommand{\BIBentryALTinterwordstretchfactor}{4}
\providecommand{\BIBentryALTinterwordspacing}{\spaceskip=\fontdimen2\font plus
\BIBentryALTinterwordstretchfactor\fontdimen3\font minus \fontdimen4\font\relax}
\providecommand{\BIBforeignlanguage}[2]{{%
\expandafter\ifx\csname l@#1\endcsname\relax
\typeout{** WARNING: IEEEtran.bst: No hyphenation pattern has been}%
\typeout{** loaded for the language `#1'. Using the pattern for}%
\typeout{** the default language instead.}%
\else
\language=\csname l@#1\endcsname
\fi
#2}}
\providecommand{\BIBdecl}{\relax}
\BIBdecl

\bibitem{mavrogiannis2023core}
C.~Mavrogiannis, F.~Baldini, A.~Wang, D.~Zhao, P.~Trautman, A.~Steinfeld, and J.~Oh, ``Core challenges of social robot navigation: A survey,'' \emph{ACM Transactions on Human-Robot Interaction}, vol.~12, no.~3, pp. 1--39, 2023.

\bibitem{korbmacher2022review}
R.~Korbmacher and A.~Tordeux, ``Review of pedestrian trajectory prediction methods: Comparing deep learning and knowledge-based approaches,'' \emph{IEEE Transactions on Intelligent Transportation Systems}, vol.~23, no.~12, pp. 24\,126--24\,144, 2022.

\bibitem{nguyen2023toward}
D.~M. Nguyen, M.~Nazeri, A.~Payandeh, A.~Datar, and X.~Xiao, ``Toward human-like social robot navigation: A large-scale, multi-modal, social human navigation dataset,'' in \emph{2023 IEEE/RSJ International Conference on Intelligent Robots and Systems (IROS)}.\hskip 1em plus 0.5em minus 0.4em\relax IEEE, 2023, pp. 7442--7447.

\bibitem{guo2025advancements}
J.~Guo, G.~Zhou, H.~Huang, and C.~Huang, ``Advancements in uav path planning: A deep reinforcement learning approach with soft actor-critic for enhanced navigation,'' \emph{Unmanned Systems}, vol.~13, no.~4, pp. 1065--1084, 2025.

\bibitem{helbing1995social}
D.~Helbing and P.~Molnar, ``Social force model for pedestrian dynamics,'' \emph{Physical Review E}, vol.~51, no.~5, p. 4282, 1995.

\bibitem{leisiazar2025adapting}
S.~Leisiazar, S.~R.~R. Rohani, E.~J. Park, A.~Lim, and M.~Chen, ``Adapting to frequent human direction changes in autonomous frontal following robots,'' \emph{IEEE Robotics and Automation Letters}, 2025.

\bibitem{lam2010human}
C.-P. Lam, C.-T. Chou, K.-H. Chiang, and L.-C. Fu, ``Human-centered robot navigation—towards a harmoniously human--robot coexisting environment,'' \emph{IEEE Transactions on Robotics}, vol.~27, no.~1, pp. 99--112, 2010.

\bibitem{antonucci2023humans}
A.~Antonucci, P.~Bevilacqua, S.~Leonardi, L.~Paolopoli, and D.~Fontanelli, ``Humans as path-finders for mobile robots using teach-by-showing navigation,'' \emph{Autonomous Robots}, vol.~47, no.~8, pp. 1255--1273, 2023.

\bibitem{algabri2020deep}
R.~Algabri and M.-T. Choi, ``Deep-learning-based indoor human following of mobile robot using color feature,'' \emph{Sensors}, vol.~20, no.~9, p. 2699, 2020.

\bibitem{gupta2016novel}
M.~Gupta, S.~Kumar, L.~Behera, and V.~K. Subramanian, ``A novel vision-based tracking algorithm for a human-following mobile robot,'' \emph{IEEE Transactions on Systems, Man, and Cybernetics: Systems}, vol.~47, no.~7, pp. 1415--1427, 2016.

\bibitem{hoy2015algorithms}
M.~Hoy, A.~S. Matveev, and A.~V. Savkin, ``Algorithms for collision-free navigation of mobile robots in complex cluttered environments: a survey,'' \emph{Robotica}, vol.~33, no.~3, pp. 463--497, 2015.

\bibitem{fox1997dynamic}
D.~Fox, W.~Burgard, and S.~Thrun, ``The dynamic window approach to collision avoidance,'' \emph{IEEE Robotics \& Automation Magazine}, vol.~4, no.~1, pp. 23--33, 1997.

\bibitem{van2011reciprocal}
J.~Van Den~Berg, S.~J. Guy, M.~Lin, and D.~Manocha, ``Reciprocal n-body collision avoidance,'' in \emph{Robotics Research: The 14th International Symposium ISRR}.\hskip 1em plus 0.5em minus 0.4em\relax Springer, 2011, pp. 3--19.

\bibitem{trautman2010unfreezing}
P.~Trautman and A.~Krause, ``Unfreezing the robot: Navigation in dense, interacting crowds,'' in \emph{2010 IEEE/RSJ International Conference on Intelligent Robots and Systems}.\hskip 1em plus 0.5em minus 0.4em\relax IEEE, 2010, pp. 797--803.

\bibitem{ryu2024integrating}
K.~Ryu and N.~Mehr, ``Integrating predictive motion uncertainties with distributionally robust risk-aware control for safe robot navigation in crowds,'' \emph{arXiv preprint arXiv:2403.05081}, 2024.

\bibitem{poddar2023crowd}
S.~Poddar, C.~Mavrogiannis, and S.~S. Srinivasa, ``From crowd motion prediction to robot navigation in crowds,'' in \emph{2023 IEEE/RSJ International Conference on Intelligent Robots and Systems (IROS)}.\hskip 1em plus 0.5em minus 0.4em\relax IEEE, 2023, pp. 6765--6772.

\bibitem{xie2023drl}
Z.~Xie and P.~Dames, ``Drl-vo: Learning to navigate through crowded dynamic scenes using velocity obstacles,'' \emph{IEEE Transactions on Robotics}, vol.~39, no.~4, pp. 2700--2719, 2023.

\bibitem{yang2023st}
Y.~Yang, J.~Jiang, J.~Zhang, J.~Huang, and M.~Gao, ``St$^{2}$: Spatial-temporal state transformer for crowd-aware autonomous navigation,'' \emph{IEEE Robotics and Automation Letters}, vol.~8, no.~2, pp. 912--919, 2023.

\bibitem{karnan2022voila}
H.~Karnan, G.~Warnell, X.~Xiao, and P.~Stone, ``Voila: Visual-observation-only imitation learning for autonomous navigation,'' in \emph{2022 International Conference on Robotics and Automation (ICRA)}.\hskip 1em plus 0.5em minus 0.4em\relax IEEE, 2022, pp. 2497--2503.

\bibitem{zhu2021deep}
K.~Zhu and T.~Zhang, ``Deep reinforcement learning based mobile robot navigation: A review,'' \emph{Tsinghua Science and Technology}, vol.~26, no.~5, pp. 674--691, 2021.

\bibitem{dugas2022flowbot}
D.~Dugas, K.~Cai, O.~Andersson, N.~Lawrance, R.~Siegwart, and J.~J. Chung, ``Flowbot: Flow-based modeling for robot navigation,'' in \emph{2022 IEEE/RSJ International Conference on Intelligent Robots and Systems (IROS)}.\hskip 1em plus 0.5em minus 0.4em\relax IEEE, 2022, pp. 8799--8805.

\bibitem{cai2023sampling}
K.~Cai, W.~Chen, D.~Dugas, R.~Siegwart, and J.~J. Chung, ``Sampling-based path planning in highly dynamic and crowded pedestrian flow,'' \emph{IEEE Transactions on Intelligent Transportation Systems}, vol.~24, no.~12, pp. 14\,732--14\,742, 2023.

\bibitem{zhu2024fast}
Y.~Zhu, A.~Rudenko, L.~Palmieri, L.~Heuer, A.~J. Lilienthal, and M.~Magnusson, ``Fast online learning of cliff-maps in changing environments,'' \emph{arXiv preprint arXiv:2410.12237}, 2024.

\bibitem{zhu2023cliff}
Y.~Zhu, A.~Rudenko, T.~P. Kucner, L.~Palmieri, K.~O. Arras, A.~J. Lilienthal, and M.~Magnusson, ``Cliff-lhmp: Using spatial dynamics patterns for long-term human motion prediction,'' in \emph{2023 IEEE/RSJ International Conference on Intelligent Robots and Systems (IROS)}.\hskip 1em plus 0.5em minus 0.4em\relax IEEE, 2023, pp. 3795--3802.

\bibitem{islam2019person}
M.~J. Islam, J.~Hong, and J.~Sattar, ``Person-following by autonomous robots: A categorical overview,'' \emph{The International Journal of Robotics Research}, vol.~38, no.~14, pp. 1581--1618, 2019.

\bibitem{scheidemann2024obstacle}
C.~Scheidemann, L.~Werner, V.~Reijgwart, A.~Cramariuc, J.~Chomarat, J.-R. Chiu, R.~Siegwart, and M.~Hutter, ``Obstacle-avoidant leader following with a quadruped robot,'' \emph{arXiv preprint arXiv:2410.00572}, 2024.

\bibitem{chen2023quadcopter}
W.-C. Chen, C.-L. Lin, Y.-Y. Chen, and H.-H. Cheng, ``Quadcopter drone for vision-based autonomous target following,'' \emph{Aerospace}, vol.~10, no.~1, p.~82, 2023.

\bibitem{lewis2009using}
M.~Lewis, H.~Wang, P.~Velagapudi, P.~Scerri, and K.~Sycara, ``Using humans as sensors in robotic search,'' in \emph{2009 12th International Conference on Information Fusion}.\hskip 1em plus 0.5em minus 0.4em\relax IEEE, 2009, pp. 1249--1256.

\bibitem{afolabi2018people}
O.~Afolabi, K.~Driggs-Campbell, R.~Dong, M.~J. Kochenderfer, and S.~S. Sastry, ``People as sensors: Imputing maps from human actions,'' in \emph{2018 IEEE/RSJ International Conference on Intelligent Robots and Systems (IROS)}.\hskip 1em plus 0.5em minus 0.4em\relax IEEE, 2018, pp. 2342--2348.

\bibitem{itkina2022multi}
M.~Itkina, Y.-J. Mun, K.~Driggs-Campbell, and M.~J. Kochenderfer, ``Multi-agent variational occlusion inference using people as sensors,'' in \emph{2022 International Conference on Robotics and Automation (ICRA)}.\hskip 1em plus 0.5em minus 0.4em\relax IEEE, 2022, pp. 4585--4591.

\bibitem{mun2023occlusion}
Y.-J. Mun, M.~Itkina, S.~Liu, and K.~Driggs-Campbell, ``Occlusion-aware crowd navigation using people as sensors,'' in \emph{2023 IEEE International Conference on Robotics and Automation (ICRA)}.\hskip 1em plus 0.5em minus 0.4em\relax IEEE, 2023, pp. 12\,031--12\,037.

\bibitem{qiu2024inferring}
T.~Qiu and D.~Fridovich-Keil, ``Inferring occluded agent behavior in dynamic games from noise corrupted observations,'' \emph{IEEE Robotics and Automation Letters}, 2024.

\bibitem{7759200}
D.~Mehta, G.~Ferrer, and E.~Olson, ``Autonomous navigation in dynamic social environments using multi-policy decision making,'' in \emph{2016 IEEE/RSJ International Conference on Intelligent Robots and Systems (IROS)}, 2016, pp. 1190--1197.

\bibitem{chen2021unified}
Y.~Chen and Y.~Lou, ``A unified multiple-motion-mode framework for socially compliant navigation in dense crowds,'' \emph{IEEE Transactions on Automation Science and Engineering}, vol.~19, no.~4, pp. 3536--3548, 2021.

\bibitem{buckeridge2023mapless}
S.~Buckeridge, P.~Carreno-Medrano, A.~Cosgun, E.~Croft, and W.~P. Chan, ``Mapless urban robot navigation by following pedestrians,'' in \emph{2023 IEEE/RSJ International Conference on Intelligent Robots and Systems (IROS)}.\hskip 1em plus 0.5em minus 0.4em\relax IEEE, 2023, pp. 6787--6792.

\bibitem{8793608}
Y.~Du, N.~J. Hetherington, C.~L. Oon, W.~P. Chan, C.~P. Quintero, E.~Croft, and H.~Machiel Van~der Loos, ``Group surfing: A pedestrian-based approach to sidewalk robot navigation,'' in \emph{2019 International Conference on Robotics and Automation (ICRA)}, 2019, pp. 6518--6524.

\bibitem{ding2019safe}
W.~Ding, L.~Zhang, J.~Chen, and S.~Shen, ``Safe trajectory generation for complex urban environments using spatio-temporal semantic corridor,'' \emph{IEEE Robotics and Automation Letters}, vol.~4, no.~3, pp. 2997--3004, 2019.

\bibitem{zhang2020trajectory}
T.~Zhang, M.~Fu, W.~Song, Y.~Yang, and M.~Wang, ``Trajectory planning based on spatio-temporal map with collision avoidance guaranteed by safety strip,'' \emph{IEEE Transactions on Intelligent Transportation Systems}, vol.~23, no.~2, pp. 1030--1043, 2020.

\bibitem{bai2025realm}
R.~Bai, S.~Yuan, K.~Li, H.~Guo, W.-Y. Yau, and L.~Xie, ``Realm: Real-time line-of-sight maintenance in multi-robot navigation with unknown obstacles,'' \emph{arXiv preprint arXiv:2502.15162}, 2025.

\bibitem{madjid2025trajectory}
N.~A. Madjid, A.~Ahmad, M.~Mebrahtu, Y.~Babaa, A.~Nasser, S.~Malik, B.~Hassan, N.~Werghi, J.~Dias, and M.~Khonji, ``Trajectory prediction for autonomous driving: Progress, limitations, and future directions,'' \emph{arXiv preprint arXiv:2503.03262}, 2025.

\bibitem{robicquet2016learning}
A.~Robicquet, A.~Sadeghian, A.~Alahi, and S.~Savarese, ``Learning social etiquette: Human trajectory understanding in crowded scenes,'' in \emph{Computer Vision--ECCV 2016: 14th European Conference, Amsterdam, The Netherlands, October 11-14, 2016, Proceedings, Part VIII 14}.\hskip 1em plus 0.5em minus 0.4em\relax Springer, 2016, pp. 549--565.

\bibitem{levine1999pace}
R.~V. Levine and A.~Norenzayan, ``The pace of life in 31 countries,'' \emph{Journal of Cross-Cultural Psychology}, vol.~30, no.~2, pp. 178--205, 1999.

\bibitem{koenig2004design}
N.~Koenig and A.~Howard, ``Design and use paradigms for gazebo, an open-source multi-robot simulator,'' in \emph{2004 IEEE/RSJ international conference on intelligent robots and systems (IROS)(IEEE Cat. No. 04CH37566)}, vol.~3.\hskip 1em plus 0.5em minus 0.4em\relax Ieee, 2004, pp. 2149--2154.

\bibitem{liu2024height}
S.~Liu, H.~Xia, F.~C. Pouria, K.~Hong, N.~Chakraborty, and K.~Driggs-Campbell, ``Height: Heterogeneous interaction graph transformer for robot navigation in crowded and constrained environments,'' \emph{arXiv preprint arXiv:2411.12150}, 2024.

\bibitem{gupta2018social}
A.~Gupta, J.~Johnson, L.~Fei-Fei, S.~Savarese, and A.~Alahi, ``Social gan: Socially acceptable trajectories with generative adversarial networks,'' in \emph{IEEE Conference on Computer Vision and Pattern Recognition (CVPR)}, no. CONF, 2018.

\bibitem{dugas2020ian}
D.~Dugas, J.~Nieto, R.~Siegwart, and J.~J. Chung, ``Ian: Multi-behavior navigation planning for robots in real, crowded environments,'' in \emph{2020 IEEE/RSJ International Conference on Intelligent Robots and Systems (IROS)}.\hskip 1em plus 0.5em minus 0.4em\relax IEEE, 2020, pp. 11\,368--11\,375.

\bibitem{che2020efficient}
Y.~Che, A.~M. Okamura, and D.~Sadigh, ``Efficient and trustworthy social navigation via explicit and implicit robot--human communication,'' \emph{IEEE Transactions on Robotics}, vol.~36, no.~3, pp. 692--707, 2020.

\bibitem{angelopoulos2022you}
G.~Angelopoulos, A.~Rossi, C.~Di~Napoli, and S.~Rossi, ``You are in my way: non-verbal social cues for legible robot navigation behaviors,'' in \emph{2022 IEEE/RSJ International Conference on Intelligent Robots and Systems (IROS)}.\hskip 1em plus 0.5em minus 0.4em\relax IEEE, 2022, pp. 657--662.

\bibitem{ravankar2018path}
A.~Ravankar, A.~A. Ravankar, Y.~Kobayashi, Y.~Hoshino, and C.-C. Peng, ``Path smoothing techniques in robot navigation: State-of-the-art, current and future challenges,'' \emph{Sensors}, vol.~18, no.~9, p. 3170, 2018.

\end{thebibliography}
